# A Comparative Analysis of Interactive Reinforcement Learning Algorithms in Warehouse Robot Grid Based Environment

Arunabh Bora
MS  Robotics & Autonomous Systems
University of Lincoln
*Student ID - 27647565*
27647565@students.lincoln.ac.uk

*Abstract*— **Warehouse robot is a highly demanding field at present. Major technology and logistics companies are investing heavily in developing these advanced systems. Training robots to operate in such a complex environment is very challenging. In most of the cases, robot requires human supervision to adapt and learn from the environment situations. Interactive reinforcement learning (IRL) is one of the training methodologies use in the field of human-computer interaction. This paper approaches a comparative study between two interactive reinforcement learning algorithms: Q-learning and SARSA. The algorithms are trained in a virtual grid-simulation-based warehouse environment. To avoid varying feedback reward biases, same-person feedback has been taken in the study.**

## I. INTRODUCTION

Autonomous Robot for Warehouse is a highly trending name now a days. Company like Amazon has already deployed over 750,000 robots to work collaboratively in the their warehouses and workstations [1]. There are multiple reasons why some of the logistic companies are willing to deploy warehouse robots. Major reasons are they are very cost-effective, highly efficient for some important tasks of warehouse like packing, picking and transporting goods. However optimizing the autonomous warehouse robots is a very challenging also rewarding task. Most of the advanced robots uses the computer vision methods or advanced level reinforcement learning methodology.

This paper introduces a simple grid-world simulation that serves as an environment for a virtual robot agent. This robot can move only to the define setup of grid world. The primary task is to demonstrate to go the goal step, and avoid other states. To demonstrate the robot agent, I have employed the Interactive Reinforcement Learning (IRL) algorithms [2]. Fundamentally, I consider the grid world to be a warehouse, and the goal state represents the robot's designated task position. Upon reaching this point from any starting location within the warehouse, the robot can then proceed with its other tasks.

There are multiple methodologies for Human Computer Interactions. Primarily, Learning from Demonstration (LfD), Interactive Machine Learning (IML) and Interactive Reinforcement Learning (IRL). In IRL specifically, the human supervisor has a clear means of influencing learning - reward. Reward can be positive (reinforcing desired behaviour) or negative (discouraging undesired behaviour).

My research question focuses on how the incorporation of human feedback in a warehouse robot, within an interactive reinforcement learning (IRL) environment loop impacts the performance of different value-based reinforcement learning algorithms.

To train the robot agent, I have employed two Reinforcement Learning algorithms to utilize the human computer interaction though IRL, first one is Q-learning and second one is SARSA. With the same person feedback reward system, I have compared the reinforcement learning algorithms behaviours, their success rate, average reward, steps taken to reach the goal state etc. To get a fix reward system, I have assigned negative and positive reward methodology. With the help of PyGame visualization, I have created an interactive environment for the robot agent and human.

The initial part of the paper covers the introduction and motivation of the study. Second part delves into the related works previously did by researchers, third part describes the theoretical background of algorithms, fourth part delineates the methodological works and processes of this study. Finally, the fifth part describes the results and analysis and sixth part is about the conclusion of the study.

## II. LITERATURE REVIEW

In the past decade, the progress of Human Computer Interaction (HCI) was incredible. Many researchers had already developed several methodologies. The paper [2] provides an extensive overview of interactive reinforcement learning (IRL) methodologies. It offers valuable insights about different IRL algorithms that utilize human feedback. These include model-based methods, model-free methods and the hybrid methods. Also, the paper explores different human feedback methodologies such as unimodal feedback (hardware delivered & natural interaction), and multi-model feedback system (speech-gesture, hardware-facial feedback system and many more).

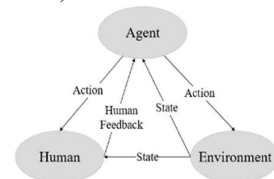

Fig(1) - Interactive Reinforcement Learning Framework [2]

Human computer feedback system varies in different scenario [2], the paper [3] examines four feedback adviser approaches – probabilistic early, importance, and mistake correcting advising within a domestic robot scenario. The paper also tells the importance of choosing appropriate advising strategy. It shows the mistake correcting advising approach achieves the better results than others.

The paper [4] presents an interactive Q-learning approach to enhance the robot autonomy. The paper also suggests an adaptive way that adjusts behaviour to match individual preferences and needs. It discusses the state-of-the-art in robot learning and experimental setups.

The paper [5] introduces a novel approach utilizing a Deep Q-Network (DQN) model to address the dispatching and routing problems in warehouse environment for robot. DQN is a value-based reinforcement learning. The authors of paper[5] compares their model with Shortest Travel Distance[STD] rule, and the DQN outperforms the STD. So, DQN can be good option for Interactive reinforcement learning algorithm in warehouse simulation.

The paper [6] shows an innovative approach in human-computer interaction through Deep Reinforcement Learning for autonomous drones, which can be use for autonomous robots also. The paper discusses total 3 DRL algorithms comparison for the task, the authors find that PPO outperforms other two algorithm DDPG and SAC. So, the authors decided to enhance the DDPG algorithm through a Human-in-the-loop-feedback. And the authors are able to achieve a great result as expectation.

## III. THEORITICAL BACKGROUND

Reinforcement Learning [2] is a framework in which agents learn to solve sequential decision-making problems by interacting with an environment to maximize cumulative rewards. RL algorithms are usually divided into three categories: policy search methods, value function methods and actor-critic methods [2].

In the reinforcement learning the Markov Decision Process (MDP) is utilize to formulate the concept. In the MDP, with states S, actions A, transition reward T, reward function R and a discount factor $\gamma$, the goal is to basically find the optimal policy $\pi^*$, which can maximize the expected future discounted reward.

The value function $V^\pi$ represents the expected future discounted reward from each state under the policy $\pi$. And the action-value function $Q^\pi$ represents the expected future discounted reward when taking a specific action in a state and then following policy $\pi$.

So mathematically, this can be shown by using Bellman Equation [7] –

$$V^\pi(s) = \sum_a \pi(s,a) Q^\pi(s,a)$$

$$Q^\pi(s,a) = \sum_{s'} T(s'|s,a) [R(s,a,s') + \gamma V^\pi(s')]$$

Interactive Reinforcement Learning [Fig(1)] is a framework where humans provides the feedback or guidance during agent learning process.

## IV. METHODOLOGY

For this research study, I have employed two Reinforcement Learning Algorithms – Q-learning and SARSA. Q-learning and SARSA (State-Action-Reward-State-Action) are both value-based reinforcement learning algorithms. They learn the value function (Q-values) associated with state-action pairs but not from the policy directly. Instead, they derive policy from the learned value function by selecting actions with the highest Q-values. To make these algorithms an Interactive Reinforcement Learning Algorithm, I have modified the action & reward system of the algorithm. Q-learning is an off-policy algorithm and SARSA is an on-policy algorithm. The primary difference between these two reinforcement learning algorithms is how they update their values, Q-learning picks the best possible action for the next step whereas SARSA chooses the actual action taken. The below figures [Fig(2) & Fig(3)] illustrate my two algorithms workflow pseudo code.

### A. Interactive Q-Learning

This algorithm [Fig(2)] is a modified version of the Q-learning algorithm. The Q-values are updated based on the human acceptance received and the maximum expected future reward.

```
1. Initialize Q-table, α, γ
2. while not end of interaction do:
    a. a_t = ε-greedy action selection for state s_t
    b. if Human accepts a_t: -
        i.  Execute a_t, observe reward r_t from
            environment, next state s_(t+1)
        ii. Update Q-value (based on agent's action
            and environment reward): Q(s_t, a_t) =
            Q(s_t, a_t) + α (r_t + γ * max_a'
            Q(s_(t+1), a') - Q(s_t, a_t) )
    c. else:
        i.  Execute a_t, observe human-provided
            reward r_human, next state s_(t+1)
        ii. Update Q-value (based on agent's action
            and environment reward): Q(s_t, a_t) =
            Q(s_t, a_t) + α (r_t + γ * max_a'
            Q(s_(t+1), a') - Q(s_t, a_t) )
    d. s_t = s_(t+1)
3. end while
```

Fig (2) – Interactive Q-Learning for learning from the Human

### B. Interactive SARSA

This algorithm [Fig(3)] is a modified version of SARSA. This approach maintain the core SARSA update while incorporating the Human feedback in a way that can influence the agent's learning without directly modifying the Q-value based on the chosen action. Because in standard SARSA the reward comes from the environment based on the executed action.

```
1. Initialize Q-table, α, γ
2. while not end of interaction do:
    a. a_t = ε-greedy action selection for state s_t
    b. if Human accepts a_t: -
        i.   Execute a_t, observe reward r_t from
             environment, next state s_(t+1)
        ii.  a_(t+1) = ε-greedy action selection for state
             s_(t+1)
        iii. Update Q-value (based on agent's action):
             Q(s_t, a_t) = Q(s_t, a_t) + α (r_t + γ *
             Q(s_(t+1), a_(t+1)) - Q(s_t, a_t) )
    c. else: -
        i.   Execute a_t, observe human-provided
             reward r_human, next state s_(t+1)
        ii.  a_(t+1) = ε-greedy action selection for state
             s_(t+1)
        iii. Update Q-value (based on agent's action):
             Q(s_t, a_t) = Q(s_t, a_t) + α (use
             environment reward+ γ * Q(s_(t+1),
             a_(t+1)) - Q(s_t, a_t) )
    d. s_t = s_(t+1)
    e. human reward
3. end while
```

Fig (3) – Interactive SARSA for learning from the Human

I have designed a grid-based simulation using the Pygame library to visualize agent performance in real-time. This function essentially constructs a warehouse environment simulation on a grid. The robot is represented by a blue circle, the goal state is a green square, and the lose states are in black color squares. Figure 4 illustrates the complete visualization process.

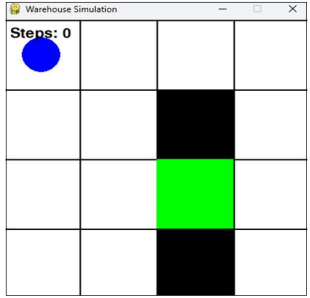

Fig (4) – PyGame Visualisation for Agent Real-time Performance Monitoring

The following table [Table(1)] represents the Hyperparameters and Parameters for both learning algorithms.

**Table (1) - Hyperparameters and Parameters Of Both Algorithms**

| Hyperparameters and Parameters | Interactive Q-Learning | Interactive SARSA |
|---|---|---|
| Grid Size | 4 | 4 |
| Cell Size | 100 | 100 |
| Number of Training Episodes | 100 | 100 |
| Maximum Steps | 120 | 120 |
| Learning Rate | 0.001 | 0.001 |
| Discount Factor | 0.89 | 0.89 |
| Epsilon | 0.97 | 0.99 |
| Epsilon_decay | 0.99 | 0.98 |
| Actions | UP,DOWN, LEFT,RIGHT | UP,DOWN, LEFT,RIGHT |
| Wining Position | (2,2) | (2,2) |
| Lose Position 1 | (1,2) | (1,2) |
| Lose Position 2 | (3,2) | (3,2) |
| Wining Reward | 10 | 10 |
| Losing Reward | -10 | -10 |

The algorithms are designed for primarily training purposes. But, the code saved the Q-table as a pickle file, which can be used for testing purposes.

## V. RESULT

Both of the learning algorithms performed well overall. In the training phase, Interactive SARSA takes more time and steps than Interactive Q-Learning. The following figure is the representation of the number of Episodes vs the number of steps required for a typical convergence.

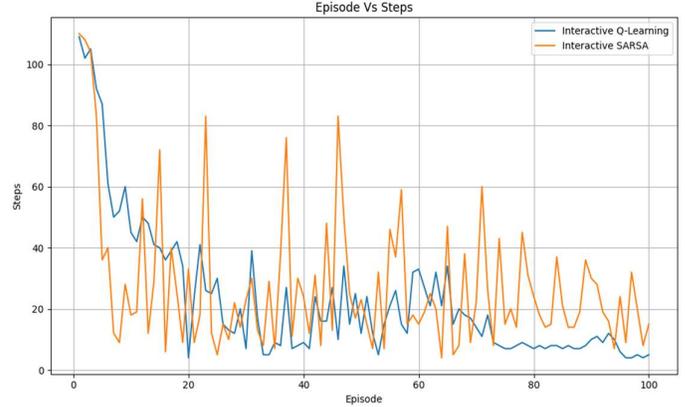

Fig(5) – Episode Vs Number of Steps Required In Training

As the algorithm is designed primarily for training purposes, it provides real-time rewards per episode during the learning phase. The reward system works as follows:

- +10: The robot achieves the winning state.
- -10: The agent enters a losing state.
- 0: The robot doesn't reach any state (win or lose) within 120 steps.

The below figure [Fig(6)] shows the average to total rewards per episode for both Interactive Q-learning and Interactive SARSA.

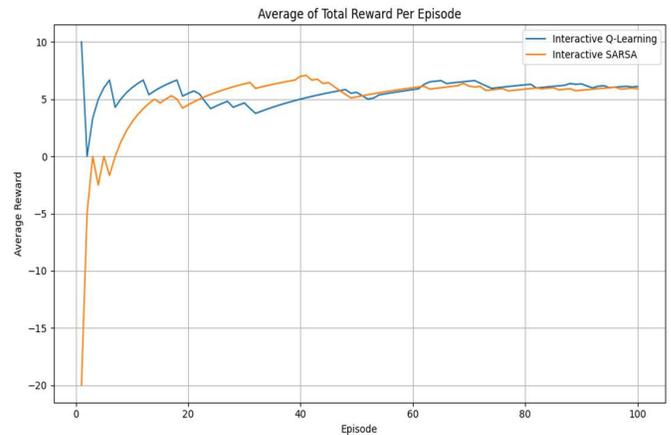

Fig (6) – Average of Total Reward per Episode

The above figure [Fig(6)] shows that the average of the total reward graph looks similar for both algorithms after 50 to 60 training episodes. This signifies that as the learning process increases the agent is able to learn the trajectory of the environment. Also, from the figure [Fig(5)] it's clearly visible that the number of steps gets reduced as the number of learning episodes increases.

After completing the training phase, the algorithm shows the Q-table for the agent. The below figure [Fig(7)] shows a breif idea of the updated Q-table for both algorithms. To draw the graph of average Q-values for each action, I have calculated the mean of the Q table elements.

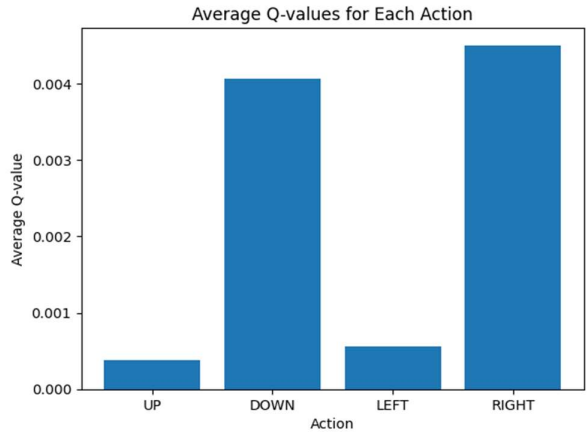

Fig (7) – Average Q-values for each action

The above figure [Fig(7)] demonstrates that the most demanding actions taken are RIGHT and DOWN.

The below table [Table (2)] is derived from the training history data.

**Table (2) – Performance Metrics of Both Algorithms**

| Performance Metrics | Interactive Q-Learning | Interactive SARSA |
|---|---|---|
| Average of Total Reward per Episode | 7.5 | 3.8 |
| Success Rate | 75% | 60% |
| Average Number of Steps per Episode | 12 | 16.13 |
| Exploration Rate | 30% | 25% |

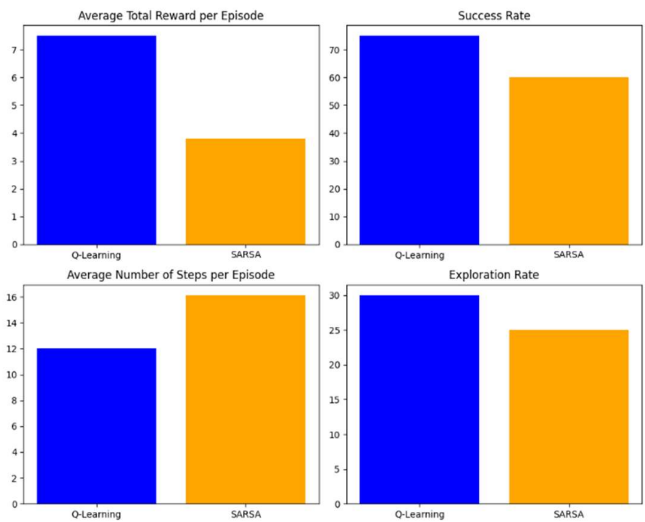

Fig (8) – Comparison Metrics of The Learning Algorithms

From the above table [Table (2)] and [Fig(8)], it is clearly visible that Interactive Q-learning performs better than Interactive SARSA.

## VI. Conclusion

Emerging fields like Robotics and AI in current days are dominating all industries. Researchers and Scientists are trying to apply these advanced approaches in all possible domains. Logistics companies are endeavouring to integrate advanced technological fields into the demanding tasks within warehouses.

In this paper, I have introduced two human-in-loop-feedback methodology compatible interactive reinforcement learning algorithms, which are responsible for training the robot agent to go to a specific position within the grid simulation-based warehouse environment. I have modified the two primary basic value-based reinforcement learning algorithms – Q-learning and SARSA. My algorithms have the capability to train the robotic agent by utilizing real-time human-provided rewards for each action undertaken.

As a result, the interaction between human-provided rewards and the agent actions, both of the algorithms have achieved an overall good performance. Interactive Q-learning has shown a success rate of 75% and the Interactive SARSA has shown a 60% success rate in achieving the goal position.

While both algorithms have performed well, they still have limitations. The primary limitation is the training time. A higher learning rate can address this, but achieving a deeper understanding of the environment requires a more balanced approach. Simply providing additional human rewards may not be the most effective solution. Alternatively, other deep reinforcement learning algorithms could perform well, algorithms such as DQN, PPO, DDPG are some better alternative approaches for Human Computer Interaction.